\documentclass[11pt,letterpaper]{article}
\usepackage{emnlp2017}
\usepackage{times}
\usepackage{latexsym}

\usepackage{url}
\usepackage{adjustbox}
\usepackage{amsmath}
\usepackage{arydshln}
\emnlpfinalcopy % Uncomment this line for the final submission

\title{Modeling Multi-Action Policy for Task-Oriented Dialogues}
\author{
Lei Shu, Hu Xu, Bing Liu\\
University of Illinois at Chicago\\
\{lshu3, hxu48, liub\}@uic.edu\\
\And
Piero Molino\\
Uber AI\\
piero@uber.com \\
}
\date{}

\begin{document}

\maketitle

\begin{abstract}
Dialogue management (DM) plays a key role in the quality of the interaction with the user in a task-oriented dialogue system.
In most existing approaches, the agent predicts only one DM policy action per turn.
This significantly limits the expressive power of the conversational agent and introduces unwanted turns of interactions that may challenge users' patience.
Longer conversations also lead to more errors and the system needs to be more robust to handle them.
In this paper,
%we aim to focus on
%novel task of predicting multiple acts per each turn to address those issues.
%In this paper, we propose a novel problem that enables the generation of multiple acts per turn (we call it multi-act) to address the above issue.
we compare the performance of several models on the task of predicting multiple acts for each turn. A novel policy model is proposed based on a recurrent cell called gated \underline{C}ontinue-\underline{A}ct-\underline{S}lots (gCAS) that overcomes the limitations of the existing models.
%We formulate a multi-act as a sequence of 
%$(\textit{continue}, \textit{act}, \textit{slot})$ 
%tuples and generate such sequence via a novel 
%recurrent cell called gated \underline{C}ontinue \underline{A}ct \underline{S}lot (gCAS).
Experimental results show that gCAS\footnote{The code is available at \url{https://leishu02.github.io/}} outperforms other approaches.\footnote{To appear in EMNLP 2019} % on a publicly available dialogue dataset.
 
\end{abstract}

\section{Introduction}
In a task-oriented dialogue system, 
%teaching a system how to respond appropriately is non-trivial. 
the dialogue manager policy module predicts actions usually in terms of 
dialogue acts and
domain specific slots.
It is a crucial component that influences the efficiency (e.g., the conciseness and smoothness) of the communication between the user and the agent.
%we implicitly make related work here.
Both supervised learning (SL)~\cite{stent2002conversation, Jason17, williams2016end, henderson2005hybrid, henderson2008hybrid} and reinforcement learning (RL) approaches~\cite{Walker2000AnAO,Young2007TheHI,Gasic2014GaussianPF,Williams2017HybridCN,Su2017SampleefficientAR} have been adopted to learn policies.
SL learns a policy to predict acts given the dialogue state. Recent work~\cite{wen2017latent,BingNAACL18} also used SL as pre-training for RL to mitigate the sample inefficiency of RL approaches and to reduce the number of interactions.
% with users in order to achieve good results.
Sequence2Sequence (Seq2Seq)~\cite{sutskever2014sequence} approaches have also been adopted in user simulators to produce user acts~\cite{Gur2018UserMF}.
These approaches typically assume that the agent can only produce one act per turn through classification.
%(? need an example with both one act--that may introduce trouble later and two acts)
%Classification format flattens dialogue acts and slots. It loses the acts dependency like 'inform (moviename)' is before 'multiple\_choice (moviename)'. Sequence-to-sequence is able to remember the act order sequence. However, it may wrongly model the same act's slots dependency. For example, in 'inform (moviename, genre)', 'moviename' and 'genre' are independent slots. Also, it is hard for sequence-to-sequence to learn out separator symbols pattern correctly like `(' after an act and `)' closing slots from limited training data.
Generating only one act per turn significantly limits what an agent can do in a turn
%and later what a user can respond within a short time frame.
%As such, a potential longer dialogue is produced and the AI agent is challenged to handle longer dialogue with acts of even high quality.
and leads to lengthy dialogues, making tracking of state and context throughout the dialogue harder.
An example in Table~\ref{tab:dialog_example} shows how the agent can produce both an \textit{inform} and a \textit{multiple\_choice} act, reducing the need for additional turns. % repetitive (which can lead to open questions that are harder for the agent to handle).

%Although multiple acts have been explored with interaction managers keeping track of the floor (who is speaking right now)~\cite[among others]{Raux2007AMA,Khouzaimi2015TurntakingPI,Hastie2013DemonstrationOT}, generating them simultaneously at each turn has been largely ignored in dialogue management and only explored with simulated data~\cite{chandramohan2010user}.
The use of multiple actions has previously been used in interaction managers that keep track of the floor (who is speaking right now)~\cite[among others]{Raux2007AMA,Khouzaimi2015TurntakingPI,Hastie2013DemonstrationOT}, but the option of generating multiple acts simultaneously at each turn for dialogue policy has been largely ignored, and only explored in simulated scenarios without real data~\cite{chandramohan2010user}.

\begin{table}
  \centering
  \resizebox{\columnwidth}{!}{
  \begin{tabular}{l|l}
%  \hline
user msg & Hi! I'm looking for good thriller.
          Are there any playing right now? \\
%\hline
%user&greeting(greeting=Hi)     \\
%acts&request(moviename;genre=thriller;starttime=right now) \\\hline    

\hline                            
agent msg & Yes, there are! The Witch, The Other Side of the Door, \\
          & and The Boy are all thrillers. Would you like to\\
          & find tickets for a showing for any of them?    \\

\hline
agent acts & inform(moviename={The Witch, The Other Side of the Door, The Boy}; \\
           & genre=thriller) multiple\_choice(moviename)    \\

\hline

\iffalse
user&    The Witch sounds scary! Can I get 4 tickets in Carbondale, Illinois for Tuesday night?    \\
acts&inform(moviename=The Witch)  request(ticket;numberofpeople=4;city=Carbondale;state=Illinois;date=Tuesday night)    \\
agent&    They are only showing The Witch at 4:40 pm.  Would that be acceptable for you?    \\
acts&inform(moviename=The Witch;starttime=4:40pm)     confirm\_question()\\\hline
user&    That'll be fine. We'll movie before dinner.    \\
acts&confirm\_answer()    inform(starttime=before dinner)    \\                
agent&    I've purchased 4 tickets for The Witch at AMC SHOWPLACE CARBONDALE 8 for 4:40 PM. Thank you \\    acts&inform(taskcomplete;numberofpeople=4;moviename=The Witch;theater=AMC SHOWPLACE CARBONDALE 8;\\&city=Carbondale;starttime=4:40PM)     thanks(closing=thank you)\\\hline    
\fi

  \end{tabular}
  }
  \caption{Dialogue example.}
  \label{tab:dialog_example}
  \vspace*{-2ex}
\end{table}

This task can be cast as a multi-label classification problem (if the sequential dependency among the acts is ignored) or as a sequence generation one as shown in Table~\ref{tab:format}.

In this paper, we introduce a novel policy model to output multiple actions per turn (called multi-act), generating a sequence of tuples and expanding agents' expressive power.
Each tuple is defined as $(\textit{continue}, \textit{act}, \textit{slots})$, where \textit{continue} indicates whether to continue or stop producing new acts, \textit{act} is an \textit{act type} (e.g., \textit{inform} or \textit{request}), and \textit{slots} is a set of slots (names) associated with the current \textit{act type}.
Correspondingly, a novel decoder (Figure~\ref{fig:casdecoder}) is proposed to produce such sequences.
Each tuple is generated by a cell called gated \underline{C}ontinue \underline{A}ct \underline{S}lots (gCAS, as in Figure~\ref{fig:cascell}), which is composed of three sequentially connected gated units handling the three components of the tuple.
This decoder can generate multi-acts in a double recurrent manner~\cite{tay2018recurrently}.
We compare this model with baseline classifiers and sequence generation models and show that it consistently outperforms them. % on the MSR challenge dataset~\cite{msr_challenge}.

%The main contributions of this work are (1) format multiple dialogue acts and slots as a sequence of (continue, act, slot) tuple; (2) propose a decoder architecture which output the tuple at each recurrent step; (3) design a novel recurrent cell gated CAS and experiment results show that gated CAS outperforms classification, sequence-to-sequence architectures; (4) we will release source code for community to explore multiple dialogue acts generation.

\begin{table*}
  \centering
  \resizebox{\linewidth}{!}{
  \begin{tabular}{l|l}
  annotation & inform(moviename={The Witch, The Other Side of the Door, The Boy}; genre=thriller)     multiple\_choice(moviename)\\
  \hline
  classification & inform+moviename, inform+genre, multiple\_choice+moviename\\\hline
  sequence & `inform' `(' `moviename' `=' `;' `genre' `=' `)' `multiple\_choice' `(' `moviename' `)' `$\langle$eos$\rangle$'\\
  \hline
  cas sequence & ($\langle$continue$\rangle$, inform, \{moviename, genre\}) ($\langle$continue$\rangle$, multiple\_choice, \{moviename\}) ($\langle$stop$\rangle$, $\langle$pad$\rangle$, \{\})\\
  \hline
  \end{tabular}
  }
  \caption{Multiple dialogue act format in different architectures.}
  \label{tab:format}
\end{table*}

% \vspace*{-1ex}
\section{Methodology}

\begin{figure}
\centering    
\includegraphics[width=\columnwidth]{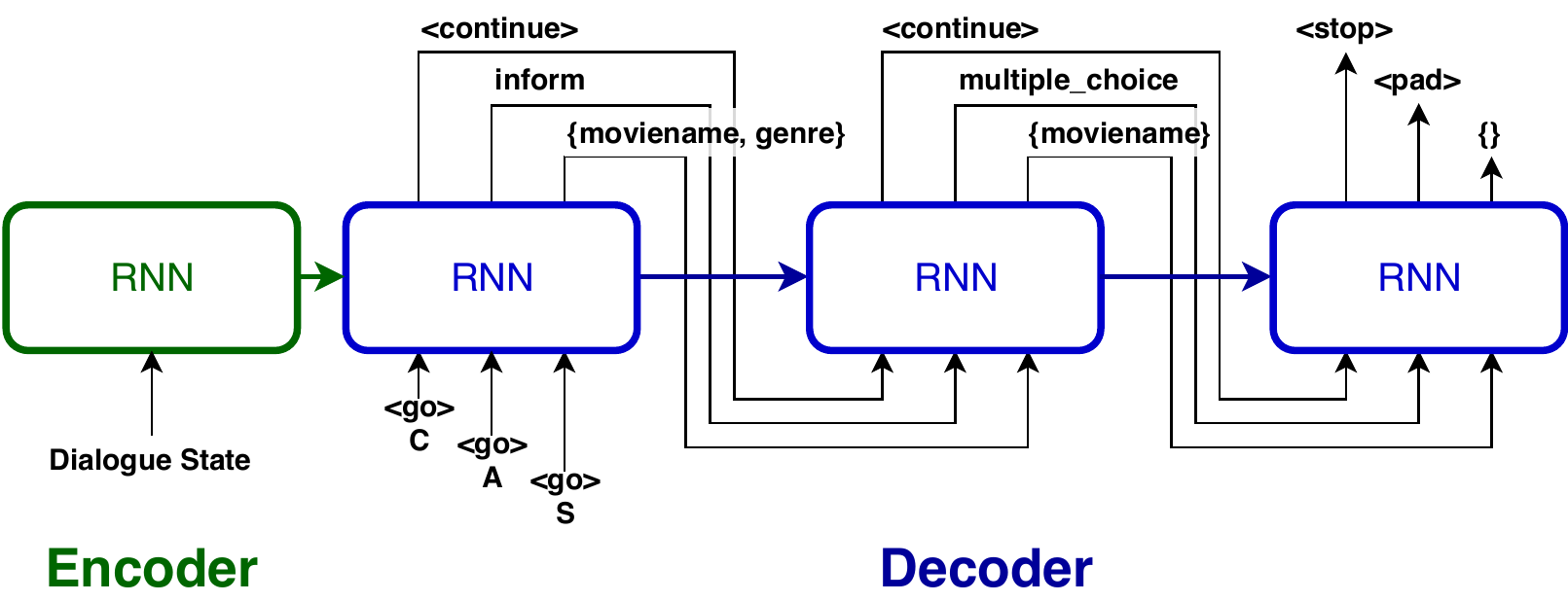}
\caption{CAS decoder: at each step, a tuple of (continue, act, slots) is produced. The KB vector $k$ regarding the queried result from knowledge base is not shown for brevity.
%The sequence decoded in this sequence is ($\langle$continue$\rangle$, inform, \{moviename, genre\}), ($\langle$continue$\rangle$, multiple\_choice, \{moviename\}), ($\langle$stop$\rangle$, $\langle$pad$\rangle$, \{\}).
}
\label{fig:casdecoder}
\vspace*{-0ex}
\end{figure}

The proposed policy network adopts an encoder-decoder architecture (Figure~\ref{fig:casdecoder}).
The input to the encoder is the current-turn dialogue state, which follows \citet{msr_challenge}'s definition.
It contains policy actions from the previous turn, user dialogue acts from the current turn, user requested slots, the user informed slots, the agent requested slots and agent proposed slots.
We treat the dialogue state as a sequence and adopt a GRU~\cite{cho2014gru} to encode it.
The encoded dialogue state is a sequence of vectors $\mathbf{E} = (e_0, \ldots, e_l)$ and the last hidden state is $h^{E}$. 
The CAS decoder recurrently generates tuples at each step.
It takes $h^{E}$ as initial hidden state $h_0$.
At each decoding step, the input contains the previous (continue, act, slots) tuple $(c_{t-1},a_{t-1},s_{t-1})$.
An additional vector $k$ containing the number of results from the knowledge base (KB) query and the current turn number is given as input.
The output of the decoder at each step is a tuple $(c, a, s)$, where $c \in \{\langle \text{continue} \rangle, \langle \text{stop} \rangle, \langle \text{pad} \rangle\}$, $a \in A$ (one act from the act set), and $s \subset S$ (a subset from the slot set).

\vspace*{-1ex}
\subsection{gCAS Cell}

\begin{figure}
\centering    
\includegraphics[width=\columnwidth]{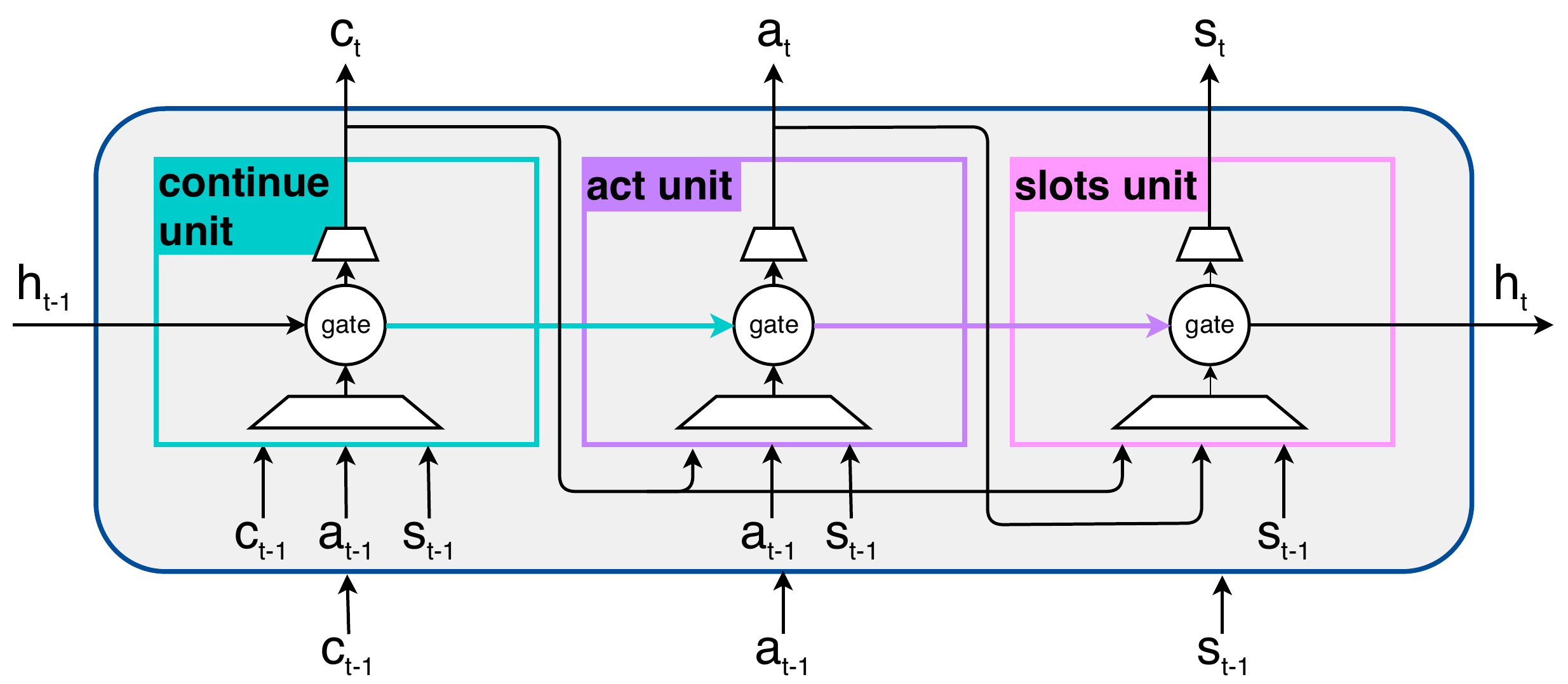}
\caption{The gated CAS recurrent cell contains three units: continue unit, act unit and slots unit. The three units use a gating mechanism and are sequentially connected. The KB vector $k$ is not shown for brevity.}
\label{fig:cascell}
\vspace*{-1ex}
\end{figure}

As shown in Figure~\ref{fig:cascell}, the gated CAS cell contains three sequentially connected units for outputting continue, act, and slots respectively.

The \textbf{Continue unit} maps the previous tuple $(c_{t-1}, a_{t-1}, s_{t-1})$ and the KB vector $k$ into $x_t^c$.
The hidden state from the previous step $h_{t-1}$ and $x_t^c$ are inputs to a $\text{GRU}^c$ unit that produces output $g_t^c$ and hidden state $h_t^c$.
Finally, $g_t^c$ is used to predict $c_t$ through a linear projection and a $\text{softmax}$.
\begin{equation}
\small
    \begin{split}
        &x_t^c = W_x^c [c_{t-1}, a_{t-1}, s_{t-1}, k] + b_x^c,\\
        &g_t^c, h_t^c = \text{GRU}^c (x_t^c, h_{t-1}), \\
        &P(c_t) = \text{softmax} (W_g^c g_t^c + b_g^c),\\
        &\mathcal{L}^c = -\sum_t \log P(c_t).\\
    \end{split}
\end{equation}

The \textbf{Act unit} maps the tuple $(c_t, a_{t-1}, s_{t-1})$ and the KB vector $k$ into $x_t^a$.
The hidden state from the continue cell $h_t^c$ and $x_t^a$ are inputs to a $\text{GRU}^a$ unit that produces output $g_t^a$ and hidden state $h_t^a$.
Finally, $g_t^a$ is used to predict $a_t$ through a linear projection and a $\text{softmax}$.
\begin{equation}
\small
    \begin{split}
        &x_t^a = W_x^a [c_t, a_{t-1}, s_{t-1}, k] + b_x^a,\\
        &g_t^a, h_t^a = \text{GRU}^a (x_t^a, h_{t}^c), \\
        &P(a_t) = \text{softmax} (W_g^a g_t^a + b_g^a),\\
        &\mathcal{L}^a = -\sum_t \log P(a_t).\\
    \end{split}
\end{equation}

The \textbf{Slots unit} maps the tuple $(c_t, a_t, s_{t-1})$ and the KB vector $k$ into $x_t^s$.
The hidden state from the act cell $h_t^a$ and $x_t^s$ are inputs to a $\text{GRU}^s$ unit that produces output $g_t^s$ and hidden state $h_t^s$.
Finally, $g_t^a$ is used to predict $s_t$ through a linear projection and a $\text{sigmoid}$. Let $z_t^i$ be the $i$-th slot's ground truth.
\begin{equation}
\small
    \begin{split}
        &x_t^s = W_x^s [c_t, a_{t}, s_{t-1}, k] + b_x^s,\\
        &g_t^s, h_t^s = \text{GRU}^s (x_t^s, h_{t}^a), \\
        &s_t = \text{sigmoid} (W_g^s g_t^s + b_g^s),\\
        &\mathcal{L}^s = -\sum_t \sum_{i = 0}^{|S|} z_t^i \log s_t^i + (1-z_t^i) \log \big(1-s_t^i\big).\\
    \end{split}
\end{equation}

The overall loss is the sum of the losses of the three units: $\mathcal{L} = \mathcal{L}^c + \mathcal{L}^a + \mathcal{L}^s$

\section{Experiments}
\begin{table}
  \centering
  \resizebox{\columnwidth}{!}{
  \begin{tabular}{c|l|l|l|l||l|l|l}
  domain    & total  & train  & valid   & test & acts & slots & pairs  \\
  \hline
  movie     & 2888   & 1445   & 433     & 1010 & 11  & 29   & 90    \\
  taxi      & 3093   & 1548   & 463     & 1082 & 11  & 23   & 63    \\
  restaurant& 4101   & 2051   & 615     & 1435 & 11  & 31   & 91    \\
  \hline
  \end{tabular}
  }
  \caption{Dataset: train, validation and test split, and the count of distinct acts, slots and act-slot pairs.}
  \label{tab:dataset}
\end{table}

\begin{table}
  \centering
  \resizebox{\columnwidth}{!}{
  \begin{tabular}{c|l|l|l|l}
  domain \& speaker & 1 act & 2 acts    & 3 acts    & 4 acts    \\
  \hline
  movie user        & 9130  & 1275      & 106       & 11        \\
  movie agent       & 5078  & 4982      & 427       & 33        \\
  \hline
  taxi user         & 10544 & 762       & 50        & 8         \\
  taxi agent        & 7855  & 3301      & 200       & 8         \\
  \hline
  restaurant user   & 12726 & 1672      & 100       & 3         \\
  restaurant agent  & 10333 & 3755      & 403       & 10        \\
  \hline
  \end{tabular}
  }
  \caption{Dialogue act counts by turn.}
  \label{tab:dataset-act}
\end{table}
\begin{table}[t]
  \centering
  \resizebox{\columnwidth}{!}{
  \begin{tabular}{l|l|l|l|l|l|l}
                & \multicolumn{3}{|c|}{Entity F$_1$}& \multicolumn{3}{|c}{Success F$_1$}\\
                \hline
                & movie         & taxi              & restaurant        & movie             & taxi          & restaurant\\
                \hline
Classification  & 34.02         & 49.71             & 28.23             & 70.41             & \textbf{84.45}& 39.97\\
\hline
Seq2Seq         & 39.95         & 63.12             & 60.21             & 77.82             & 75.09         & 55.70\\
\hline
Copy Seq2Seq    & 28.04         & 62.95             & 59.14             & 77.59             & 74.58         & 58.74\\
\hline
CAS             & 48.02         & 59.16             & 54.70             & 76.81             & 78.89         & 65.18\\
\hline\hline
gCAS            & \textbf{50.86}& \textbf{64.00}    & \textbf{60.35}    & \textbf{77.95}    & 81.17         & \textbf{71.52}\\
\hline
    \end{tabular}
  }
  \caption{Entity F$_1$ and Success F$_1$ at dialogue level.}
  \label{tab:result-task}
\end{table}

\begin{table*}
  \centering
  \resizebox{\textwidth}{!}{
  \begin{tabular}{l|c|c|c|c|c|c|c|c|c|c|c|c|c|c|c|c|c|c}
  & \multicolumn{9}{c|}{Act} & \multicolumn{9}{|c}{Frame} \\\hline
   & \multicolumn{3}{c|}{movie} & \multicolumn{3}{c|}{taxi} &  \multicolumn{3}{c}{restaurant}& \multicolumn{3}{|c|}{movie} & \multicolumn{3}{c|}{taxi} &  \multicolumn{3}{c}{restaurant}\\\hline
  method & $\mathcal{P}$&$\mathcal{R}$&$\mathcal{F}_1$ & $\mathcal{P}$&$\mathcal{R}$&$\mathcal{F}_1$&$\mathcal{P}$&$\mathcal{R}$& $\mathcal{F}_1$& $\mathcal{P}$&$\mathcal{R}$&$\mathcal{F}_1$ & $\mathcal{P}$&$\mathcal{R}$&$\mathcal{F}_1$&$\mathcal{P}$&$\mathcal{R}$& $\mathcal{F}_1$\\
  \hline
classification  &84.19  &50.24  &62.93          &92.20  &55.48  &69.27          &79.71  &33.94  &47.60          &63.91  &18.39  &28.56          &65.87  &44.31  &52.98          &49.63  &12.32  &19.74\\
\hline
Seq2Seq         &73.44  &73.62  &73.53          &77.52  &69.29  &73.17          &65.66  &66.01  &65.83          &42.88  &24.81  &31.43          &57.12  &50.32  &53.51          &39.97  &25.40  &31.06\\
\hline
Copy Seq2Seq    &67.56  &73.61  &70.46          &73.99  &69.21  &71.52          &64.93  &65.69  &65.31          &41.90  &23.12  &29.80          &51.66  &50.23  &50.93          &36.96  &27.22  &31.35  \\
\hline
CAS             &70.46  &76.08  &73.16          &79.85  &72.54  &76.02          &65.40  &72.43  &68.73          &43.12  &31.60  &36.47          &51.66  &54.29  &52.94          &33.72  &25.45  &29.01 \\
\hline\hline
gCAS            &73.08  &75.78  &\textbf{74.41} &79.47  &75.39  &\textbf{77.37} &68.30  &74.39  &\textbf{71.22} &42.24  &35.50  &\textbf{38.58} &53.77  &56.24  &\textbf{54.98} &36.86  &32.41  &\textbf{34.49} \\
\hline

  \end{tabular}
  }
  \caption{Precision ($\mathcal{P}$), Recall ($\mathcal{R}$) and F$_1$score ($\mathcal{F}_1$) of turn-level acts and frames.}
  \label{tab:result-prf}
\end{table*}

\begin{table*}
  \centering
  \resizebox{\textwidth}{!}{
  \begin{tabular}{l|l|l}
                    & example 1                                 & example 2 \\
                    \hline
  groundtruth       & request(date; starttime)                  & inform(restaurantname=; starttime =) multiple\_choice(restaurantname) \\
  \hline
  classification    & request+date                              & []    \\
  Seq2Seq           & `request' `(' `date' `;' `starttime' `)'  &  `inform' `(' `restaurantname' `=' `)' `multiple\_choice' `=' `restaurantname' `)'\\
  Copy Seq2Seq      & `request' `(' `date' `=' `)'              & `inform' `(' `restaurantname' `=' `;' `;', `;', `=',  `;' `starttime' `=' `)' \\
  CAS               & request \{\}                              & inform \{restaurantname\} \\
  gCAS              & request \{date; starttime\}               & inform \{restaurantname\} multiple\_choice\{restaurantname\}  \\
  \hline
    \end{tabular}
  }
  \caption{Examples of predicted dialogue acts in the restaurant domain.}
  \label{tab:casestudy}
\end{table*}

\begin{table*}
  \centering
  \resizebox{\textwidth}{!}{
  \begin{tabular}{l|c|c|c|c|c|c|c|c|c|c|c|c|c|c|c|c|c|c}
  & \multicolumn{9}{c|}{All Slots} & \multicolumn{9}{|c}{Non-critical slots} \\\hline
   & \multicolumn{3}{c|}{movie} & \multicolumn{3}{c|}{taxi} &  \multicolumn{3}{c}{restaurant}& \multicolumn{3}{|c|}{movie} & \multicolumn{3}{c|}{taxi} &  \multicolumn{3}{c}{restaurant}\\\hline
  method & $\mathcal{P}$&$\mathcal{R}$&$\mathcal{F}_1$ & $\mathcal{P}$&$\mathcal{R}$&$\mathcal{F}_1$&$\mathcal{P}$&$\mathcal{R}$& $\mathcal{F}_1$& $\mathcal{P}$&$\mathcal{R}$&$\mathcal{F}_1$ & $\mathcal{P}$&$\mathcal{R}$&$\mathcal{F}_1$&$\mathcal{P}$&$\mathcal{R}$& $\mathcal{F}_1$\\
  \hline
classification &67.90&21.48&32.64&73.52&72.66&73.08&45.16&12.71&19.84 &62.98&13.39&22.08&43.91&60.03&50.72&33.61&11.15&16.75\\
\hline
Seq2Seq   &53.25&29.54&38.00&64.09&74.32&68.83&42.36&17.73&25.00
&47.90&13.95&21.61&64.15&48.45&55.20&35.28&12.95&18.94\\     
\hline
Copy Seq2Seq    &52.78&28.43&36.95&56.92&74.06&64.37&38.15&22.38&28.21
&40.45&12.48&19.07&45.95&55.46&50.26&34.90&19.11&24.70\\
\hline
CAS  &63.61&33.16&43.59&61.90&80.39&69.94&51.12&22.57&31.31
&56.21&26.96&36.44&43.03&68.03&52.72&37.31&15.87&22.27\\
\hline\hline
gCAS  &54.75&38.70&45.35&62.31&79.76&69.96&44.20&29.65&35.49       
&48.23&36.68&41.67&44.35&62.15&51.77&31.26&29.60&30.41\\
\hline

  \end{tabular}
  }
  \caption{$\mathcal{P}$, $\mathcal{R}$ and $\mathcal{F}_1$ of turn-level \textit{inform} all slots and non-critical slots.}
  \label{tab:result-inform-all-prf}
\end{table*}

\begin{table}
  \centering
  \resizebox{\columnwidth}{!}{
  \begin{tabular}{l|c|c|c|c|c|c|c|c|c}
  & \multicolumn{9}{c}{Critical Slots} \\\hline
   & \multicolumn{3}{c|}{movie} & \multicolumn{3}{c|}{taxi} &  \multicolumn{3}{c}{restaurant}\\\hline
  method & $\mathcal{P}$&$\mathcal{R}$&$\mathcal{F}_1$ & $\mathcal{P}$&$\mathcal{R}$&$\mathcal{F}_1$&$\mathcal{P}$&$\mathcal{R}$& $\mathcal{F}_1$\\\hline
Classification&70.29&29.13&41.19&85.18&75.90&80.27&55.66&13.76&22.07\\\hline
Seq2Seq&55.08&44.26&49.08&64.08&80.97&71.54&46.24&20.97&28.86\\\hline
Copy Seq2Seq&57.54&43.49&49.54&59.49&78.83&67.81&40.11&24.59&30.49\\\hline
CAS&69.59&39.02&50.00&68.15&83.57&75.08&59.93&27.10&37.32\\\hline
gCAS&61.89&40.62&49.04&67.48&84.28&74.95&61.35&29.69&40.01\\\hline
  \end{tabular}
  }
  \caption{$\mathcal{P}$, $\mathcal{R}$ and $\mathcal{F}_1$ of turn-level \textit{inform} critical slots.}
  \label{tab:result-inform-critical-prf}
\end{table}

%\subsection{Dataset and Preprocessing}
The experiment dataset comes from Microsoft Research (MSR) \footnote{\url{https://github.com/xiul-msr/e2e_dialog_challenge}}. It contains three domains: movie, taxi, and restaurant.
The total count of dialogues per domain and train/valid/test split is reported in Table~\ref{tab:dataset}.
At every turn both user and agent acts are annotated, we use only the agent side as targets in our experiment. The acts are ordered in the dataset (each output sentence aligns with one act).
The size of the sets of acts, slots, and act-slot pairs are also listed in Table~\ref{tab:dataset}.
Table~\ref{tab:dataset-act} shows the count of turns with multiple act annotations, which amounts to 23\% of the dataset.
We use MSR's dialogue management code and knowledge base to obtain the state at each turn and use it as input to every model.
%As the dialogue management code may not always produce the correct state, and this can impact negatively act prediction, so the reported results should be considered a conservative estimation of models' performance in presence of correct states.

\subsection{Evaluation Metrics}
We evaluate the performance at the act, frame and task completion level.
%\footnote{MSR's user simulator does not support multi-act, which hinders interactive evaluation.}.
For a frame to be correct, both the act and all the slots should match the ground truth.
We report precision, recall, F$_1$ score of turn-level acts and frames.
For task completion evaluation, \textbf{Entity F$_1$} score and \textbf{Success F$_1$} score ~\cite{leisequicity} are reported.
The Entity F$_1$ score, differently from the entity match rate in state tracking, compares the slots requested by the agent with the slots the user informed about and that were used to perform the KB query.
We use it to measure agent performance in requesting information.
%The Entity F$_1$ evaluates how an agent requests the user informing an entity's constraints at dialogue level.
%It mainly shows the performance of the agent's `request', `multiple\_choice' acts quality.
%It is different from the entity match rate in state tracking.
The Success F$_1$ score compares the slots provided by the agent with the slots requested by the user.
We use it to measure the agent performance in providing information.
%The Success F$_1$ score evaluates how an agent responds to the user's requests at the dialogue level.
%It shows the performance of the agent's `inform' act quality.

%\subsection{Benchmarks}
\textbf{Critical slots and Non-critical slots}: 
By `non-critical', we mean slots that the user informs the system about by providing their values and thus it is not critical for the system to provide them in the output. %as it would just repeat what the user uttered. 
Table 1 shows an example, with the genre slot provided by the user and the system repeating it in its answer. Critical slots refers to slots that the system must provide like “moviename” in the Table 1 example. Although non-critical slots do not impact task completion directly, they may influence the output quality by enriching the dialogue state and helping users understand the system's utterance correctly.
Furthermore, given the same dialog state,
utterances offering non-critical slots or not offering them can both be present in the dataset, as they are optional. This makes the prediction of those slots more challenging for the system.
To provide a more detailed analysis, we report the precision, recall, F$_1$ score of turn-level for all slots, critical slots and non-critical slots of the \textit{inform} act.

\subsection{Baseline}
We compare five methods on the multi-act task.

\textbf{Classification} replicates the MSR challenge~\cite{msr_challenge} policy network architecture: two  fully connected layers.
We replace the last activation from $\text{softmax}$ to $\text{sigmoid}$ in order to predict probabilities for each act-slot pair. It is equivalent to binary classification for each act-slot pair and the loss is the sum of the binary cross-entropy of all of them.

\textbf{Seq2Seq}~\cite{sutskever2014sequence} encodes the dialogue state as a sequence, and decodes agent acts as a sequence with attention~\cite{bahdanau2014neural}. 

\textbf{Copy Seq2Seq}~\cite{gu2016incorporating} adds a copy mechanism to Seq2Seq, which allows copying words from the encoder input.

\textbf{CAS} adopts a single GRU~\cite{cho2014gru} for decoding and uses three different fully connected layers for mapping the output of the GRU to continue, act and slots. For each step in the sequence of CAS tuples, given the output of the GRU, continue, act and slot predictions are obtained by separate heads, each with one fully connected layer. The hidden state of the GRU and the predictions at the previous step are passed to the cell at the next step connecting them sequentially.

\textbf{gCAS} uses our proposed recurrent cell which contains separate continue, act and slots unit that are sequentially connected.
%At each decoder step, it outputs continue, act and slots at the same time.

The classification architecture has two fully connected layers of size 128, and the remaining models have a hidden size of 64 and a teacher-forcing rate of 0.5.
Seq2Seq and Copy Seq2Seq use a beam search with beam size 10 during inference. CAS and gCAS do not adopt a beam search since their inference steps are much less than Seq2Seq methods.
All models use Adam optimizer~\cite{kingma2014adam} with a learning rate of 0.001. 
%No dropout nor regularization is applied for purely comparing architecture performance.

\subsection{Result and Error Analysis}
As shown in Table~\ref{tab:result-task}, gCAS outperforms all other methods on Entity F$_1$ in all three domains.
Compared to Seq2Seq, the performance advantage of gCAS in the taxi and restaurant domains is small, while it is more evident in the movie domain.
%The reason is that in the restaurant and taxi domain a single KB result is returned most of the time after requesting date and time, so there is no need to use multiple acts to inform about it and request a choice, which is much more common in the movie domain.
%The reason is that in the movie domain the user usually does not know which movies are available, so the agent needs to inform about them and ask for a multiple choice, which requires multiple acts, while in the taxi and restaurant domains multiple acts are less common.
The reason is that in the movie domain the proportion of turns with multiple acts is higher (52\%), while in the other two domains it is lower (30\%).
%The reason is that taxi and restaurant has  `request' as single-act at each turn.
%On the other hand, in the movie domain, `multiple\_choice' and `request' usually appear as the second act.
%Generating these longer sequences is harder for Seq2Seq.
%Such a problem does not happen in CAS and gCAS as they only have to produce two tuples to obtain the correct prediction.
gCAS also outperforms all other models in terms of Success F$_1$ in the movie and restaurant domain but is outperformed by the classification model in the taxi domain.
The reason is that in the taxi domain, the agent usually informs the user at the last turn, while in all previous turns the agent usually requests information from the user.
It is easy for the classification model to overfit this pattern.
The advantage of gCAS in the restaurant domain is much more evident: the agent's \textit{inform} act usually has multiple slots (see example 2 in Table~\ref{tab:casestudy}) and
this makes classification and sequence generation harder, but gCAS multi-label slots decoder handles it easily.

Table~\ref{tab:result-prf} shows the turn-level acts and frame prediction performance.
CAS and gCAS outperform all other models in acts prediction in terms of F$_1$ score.
The main reason is that CAS and gCAS output a tuple at each recurrent step, which makes for shorter sequences that are easier to generate compared to the long sequences of Seq2Seq (example 2 in Table \ref{tab:casestudy}).
%The increased amount of decoding steps makes it harder for Seq2Seq to generate multiple acts.
The classification method has a good precision score, but a lower recall score, suggesting it has problems making granular decisions (example 2 in Table \ref{tab:casestudy}).
At the frame level, gCAS still outperforms all other methods.
The performance difference between CAS and gCAS on frames becomes much more evident, suggesting that gCAS is more capable of predicting slots that are consistent with the act.
This finding is also consistent with their Entity F$_1$ and Success F$_1$ performance.

However, gCAS's act-slot pair performance is far from perfect.
The most common failure case is on non-critical slots (like `genre' in the example in Table~\ref{tab:format}): gCAS does not predict them, while it predicts the critical ones (like `moviename' in the example in Table~\ref{tab:format}).
%This suggests that the current input provided to the encoder is not rich enough, providing the previous user’s utterance semantic frames or directly user’s utterance text as additional input could mitigate this problem.

Table~\ref{tab:casestudy} shows predictions of all methods from two emblematic examples.
Example 1 is a frequent single-act multi-slots agent act.
Example 2 is a complex multi-act example. 
The baseline classification method can predict frequent pairs in the dataset, but cannot predict any act in the complex example. The generated sequences of Copy Seq2Seq and Seq2Seq show that both models struggle in following the syntax.
CAS cannot predict slots correctly even if the act is common in the dataset.
gCAS returns a correct prediction for Example 1, but for Example 2 gCAS cannot predict `starttime', which is a non-critical slot.

 Tables \ref{tab:result-inform-all-prf} and \ref{tab:result-inform-critical-prf} show the results of all slots, critical slots and non-critical slots under the \textit{inform} act.
gCAS performs better than the other methods on all slots in the movie and restaurant domains. The reason why classification performs the best here in the taxi domain is the same as the Success F$_1$. In the taxi domain, the agent usually informs the user at the last turn. The non-critical slots are also repeated frequently in the taxi domain, which makes their prediction easier. 
gCAS's performance is close to other methods on critical-slots. The reason is that the \textit{inform} act is mostly the first act in multi-act and critical slots are usually frequent in the data. All methods can predict them well.

In the movie and restaurant domains, the \textit{inform} act usually appears during the dialogue and there are many optional non-critical slots that can appear (see Table \ref{tab:dataset}, movie and restaurant domains have more slots and pairs than the taxi domain). gCAS can better predict the non-critical slots than other methods. However, the overall performance on non-critical slots is much worse than critical slots since their appearances are optional and inconsistent in the data.

\section{Conclusion and Future Work}
In this paper, we introduced a multi-act dialogue policy model motivated by the need for a richer interaction between users and conversation agents.
We studied classification and sequence generation methods for this task, and proposed a novel recurrent cell, gated CAS, which allows the decoder to output a tuple at each step.~Experimental results showed that gCAS is the best performing model for multi-act prediction.~The CAS decoder and the gCAS cell can also be used in a user simulator and gCAS can be applied in the encoder.
A few directions for improvement have also been identified: 1) improving the performance on non-critical slots, 2) tuning the decoder with RL, 3) text generation from gCAS.
We leave them as future work.

\section*{Acknowledgments}
We would like to express our special thanks to Alexandros Papangelis and Gokhan Tur for their support and contribution. 
We also would like to thank Xiujun Li for his help on dataset preparation and Jane Hung for her valuable comments. Bing Liu is partially supported by the NSF grant IIS-1910424 and a research gift from Northrop Grumman. 

\bibliography{emnlp2019}

\begin{thebibliography}{}
\expandafter\ifx\csname natexlab\endcsname\relax\def\natexlab#1{#1}\fi

\bibitem[{Bahdanau et~al.(2015)Bahdanau, Cho, and Bengio}]{bahdanau2014neural}
Dzmitry Bahdanau, Kyunghyun Cho, and Yoshua Bengio. 2015.
\newblock Neural machine translation by jointly learning to align and
  translate.
\newblock In {\em International Conference on Learning Representations, San
  Diego, California, USA\/}.

\bibitem[{Chandramohan and Pietquin(2010)}]{chandramohan2010user}
Senthilkumar Chandramohan and Olivier Pietquin. 2010.
\newblock User and noise adaptive dialogue management using hybrid system
  actions.
\newblock In {\em Spoken Dialogue Systems for Ambient Environments\/},
  Springer, pages 13--24.

\bibitem[{Cho et~al.(2014)Cho, van Merrienboer, G{\"{u}}l{\c{c}}ehre, Bahdanau,
  Bougares, Schwenk, and Bengio}]{cho2014gru}
Kyunghyun Cho, Bart van Merrienboer, {\c{C}}aglar G{\"{u}}l{\c{c}}ehre, Dzmitry
  Bahdanau, Fethi Bougares, Holger Schwenk, and Yoshua Bengio. 2014.
\newblock Learning phrase representations using {RNN} encoder-decoder for
  statistical machine translation.
\newblock In {\em {EMNLP}\/}. {ACL}, pages 1724--1734.

\bibitem[{Gasic and Young(2014)}]{Gasic2014GaussianPF}
Milica Gasic and Steve~J. Young. 2014.
\newblock Gaussian processes for pomdp-based dialogue manager optimization.
\newblock {\em IEEE/ACM Transactions on Audio, Speech, and Language
  Processing\/} 22:28--40.

\bibitem[{Gu et~al.(2016)Gu, Lu, Li, and Li}]{gu2016incorporating}
Jiatao Gu, Zhengdong Lu, Hang Li, and Victor O.~K. Li. 2016.
\newblock Incorporating copying mechanism in sequence-to-sequence learning.
\newblock In {\em {ACL} {(1)}\/}. The Association for Computer Linguistics.

\bibitem[{Gur et~al.(2018)Gur, Hakkani-Tur, Tur, and Shah}]{Gur2018UserMF}
Izzeddin Gur, Dilek~Z. Hakkani-Tur, Gokhan Tur, and Pararth Shah. 2018.
\newblock User modeling for task oriented dialogues.
\newblock {\em 2018 IEEE Spoken Language Technology Workshop (SLT)\/} pages
  900--906.

\bibitem[{Hastie et~al.(2013)Hastie, Aufaure, Alexopoulos, Cuay{\'a}huitl,
  Dethlefs, Gasic, Henderson, Lemon, Liu, Mika, Mustapha, Rieser, Thomson,
  Tsiakoulis, and Vanrompay}]{Hastie2013DemonstrationOT}
Helen~F. Hastie, Marie-Aude Aufaure, Panos Alexopoulos, Heriberto
  Cuay{\'a}huitl, Nina Dethlefs, Milica Gasic, James Henderson, Oliver Lemon,
  Xingkun Liu, Peter Mika, Nesrine~Ben Mustapha, Verena Rieser, Blaise Thomson,
  Pirros Tsiakoulis, and Yves Vanrompay. 2013.
\newblock Demonstration of the parlance system: a data-driven incremental,
  spoken dialogue system for interactive search.
\newblock In {\em SIGDIAL Conference\/}.

\bibitem[{Henderson et~al.(2005)Henderson, Lemon, and
  Georgila}]{henderson2005hybrid}
James Henderson, Oliver Lemon, and Kallirroi Georgila. 2005.
\newblock Hybrid reinforcement/supervised learning for dialogue policies from
  communicator data.
\newblock In {\em IJCAI workshop on knowledge and reasoning in practical
  dialogue systems\/}. Citeseer, pages 68--75.

\bibitem[{Henderson et~al.(2008)Henderson, Lemon, and
  Georgila}]{henderson2008hybrid}
James Henderson, Oliver Lemon, and Kallirroi Georgila. 2008.
\newblock Hybrid reinforcement/supervised learning of dialogue policies from
  fixed data sets.
\newblock {\em Computational Linguistics\/} 34(4):487--511.

\bibitem[{Khouzaimi et~al.(2015)Khouzaimi, Laroche, and
  Lef{\`e}vre}]{Khouzaimi2015TurntakingPI}
Hatim Khouzaimi, Romain Laroche, and Fabrice Lef{\`e}vre. 2015.
\newblock Turn-taking phenomena in incremental dialogue systems.
\newblock In {\em EMNLP\/}.

\bibitem[{Kingma and Ba(2015)}]{kingma2014adam}
Diederik~P Kingma and Jimmy Ba. 2015.
\newblock Adam: A method for stochastic optimization.
\newblock In {\em International Conference on Learning Representations, San
  Diego, California, USA\/}.

\bibitem[{Lei et~al.(2018)Lei, Jin, Kan, Ren, He, and Yin}]{leisequicity}
Wenqiang Lei, Xisen Jin, Min-Yen Kan, Zhaochun Ren, Xiangnan He, and Dawei Yin.
  2018.
\newblock Sequicity: Simplifying task-oriented dialogue systems with single
  sequence-to-sequence architectures.
\newblock In {\em ACL\/}.

\bibitem[{Li et~al.(2018)Li, Panda, Liu, and Gao}]{msr_challenge}
Xiujun Li, Sarah Panda, Jingjing Liu, and Jianfeng Gao. 2018.
\newblock Microsoft dialogue challenge: Building end-to-end task-completion
  dialogue systems.
\newblock volume abs/1807.11125.

\bibitem[{Liu and Lane(2018)}]{BingNAACL18}
Bing Liu and Ian Lane. 2018.
\newblock End-to-end learning of task-oriented dialogs.
\newblock In {\em Proceedings of the {NAACL-HLT}\/}.

\bibitem[{Raux and Esk{\'e}nazi(2007)}]{Raux2007AMA}
Antoine Raux and Maxine Esk{\'e}nazi. 2007.
\newblock A multi-layer architecture for semi-synchronous event-driven dialogue
  management.
\newblock {\em 2007 IEEE Workshop on Automatic Speech Recognition and
  Understanding (ASRU)\/} pages 514--519.

\bibitem[{Stent(2002)}]{stent2002conversation}
Amanda~J Stent. 2002.
\newblock A conversation acts model for generating spoken dialogue
  contributions.
\newblock {\em Computer Speech \& Language\/} 16(3-4):313--352.

\bibitem[{Su et~al.(2017)Su, Budzianowski, Ultes, Gasic, and
  Young}]{Su2017SampleefficientAR}
Pei{-}Hao Su, Pawel Budzianowski, Stefan Ultes, Milica Gasic, and Steve~J.
  Young. 2017.
\newblock Sample-efficient actor-critic reinforcement learning with supervised
  data for dialogue management.
\newblock In Kristiina Jokinen, Manfred Stede, David DeVault, and Annie Louis,
  editors, {\em Proceedings of the 18th Annual SIGdial Meeting on Discourse and
  Dialogue, Saarbr{\"{u}}cken, Germany, August 15-17, 2017\/}. Association for
  Computational Linguistics, pages 147--157.

\bibitem[{Sutskever et~al.(2014)Sutskever, Vinyals, and
  Le}]{sutskever2014sequence}
Ilya Sutskever, Oriol Vinyals, and Quoc~V. Le. 2014.
\newblock Sequence to sequence learning with neural networks.
\newblock In {\em {NIPS}\/}. pages 3104--3112.

\bibitem[{Tay et~al.(2018)Tay, Luu, and Hui}]{tay2018recurrently}
Yi~Tay, Anh~Tuan Luu, and Siu~Cheung Hui. 2018.
\newblock Recurrently controlled recurrent networks.
\newblock In {\em Advances in Neural Information Processing Systems\/}. pages
  4736--4748.

\bibitem[{Walker(2000)}]{Walker2000AnAO}
Marilyn~A. Walker. 2000.
\newblock An application of reinforcement learning to dialogue strategy
  selection in a spoken dialogue system for email.
\newblock {\em J. Artif. Intell. Res.\/} 12:387--416.

\bibitem[{Wen et~al.(2017)Wen, Miao, Blunsom, and Young}]{wen2017latent}
Tsung{-}Hsien Wen, Yishu Miao, Phil Blunsom, and Steve~J. Young. 2017.
\newblock Latent intention dialogue models.
\newblock In {\em {ICML}\/}. {PMLR}, volume~70 of {\em Proceedings of Machine
  Learning Research\/}, pages 3732--3741.

\bibitem[{Williams et~al.(2017{\natexlab{a}})Williams, Asadi, and
  Zweig}]{Jason17}
Jason~D. Williams, Kavosh Asadi, and Geoffrey Zweig. 2017{\natexlab{a}}.
\newblock Hybrid code networks: practical and efficient end-to-end dialog
  control with supervised and reinforcement learning.
\newblock In {\em {ACL} {(1)}\/}. Association for Computational Linguistics,
  pages 665--677.

\bibitem[{Williams et~al.(2017{\natexlab{b}})Williams, Asadi, and
  Zweig}]{Williams2017HybridCN}
Jason~D. Williams, Kavosh Asadi, and Geoffrey Zweig. 2017{\natexlab{b}}.
\newblock Hybrid code networks: practical and efficient end-to-end dialog
  control with supervised and reinforcement learning.
\newblock In {\em ACL\/}.

\bibitem[{Williams and Zweig(2016)}]{williams2016end}
Jason~D Williams and Geoffrey Zweig. 2016.
\newblock End-to-end lstm-based dialog control optimized with supervised and
  reinforcement learning.
\newblock {\em arXiv preprint arXiv:1606.01269\/} .

\bibitem[{Young et~al.(2007)Young, Schatzmann, Weilhammer, and
  Ye}]{Young2007TheHI}
Steve~J. Young, Jost Schatzmann, Karl Weilhammer, and Hui Ye. 2007.
\newblock The hidden information state approach to dialog management.
\newblock {\em 2007 IEEE International Conference on Acoustics, Speech and
  Signal Processing - ICASSP '07\/} 4:IV--149--IV--152.

\end{thebibliography}
\bibliographystyle{emnlp_natbib}

\end{document}